\documentclass[sigconf]{acmart}

\usepackage{bookmark}
\usepackage{amsfonts}
\usepackage{enumitem}
\usepackage{multirow}
\setlist[itemize]{leftmargin=*}
\setlist[enumerate]{leftmargin=*}

\graphicspath{ {./figures/} }

\AtBeginDocument{%
  \providecommand\BibTeX{{%
    \normalfont B\kern-0.5em{\scshape i\kern-0.25em b}\kern-0.8em\TeX}}}

\setcopyright{acmcopyright}
\copyrightyear{2021}
\acmYear{2021}
\acmDOI{}

\acmConference[ICAIF '21]{ICAIF '21}{November 3--5, 2021}{Online}
\acmBooktitle{Proceedings of ACM International
Conference on AI in Finance, Nov 2021}
\acmPrice{}
\acmISBN{}

\begin{document}

\settopmatter{printfolios=true}
\title{Active learning for imbalanced data under cold start}

\author{Ricardo Barata}
\email{ricardo.barata@feedzai.com}
\affiliation{%
  \country{Feedzai}
}

\author{Miguel Leite}
\authornote{Work developed while employed at Feedzai.}
\email{miguel.leite@feedzai.com}
\affiliation{%
   \country{Feedzai}
}

\author{Ricardo Pacheco}
\authornotemark[1]
\email{rjgpacheco@gmail.com}
\affiliation{%
   \country{Feedzai}
}

\author{Marco O. P. Sampaio}
\email{marco.sampaio@feedzai.com}
\affiliation{%
   \country{Feedzai}
}

\author{Jo\~ao Tiago Ascens\~ao}
\email{joao.ascensao@feedzai.com}
\affiliation{%
   \country{Feedzai}
}

\author{Pedro Bizarro}
\email{pedro.bizarro@feedzai.com}
\affiliation{%
   \country{Feedzai}
}

\renewcommand{\shortauthors}{Barata, et al.}

\begin{abstract}
Modern systems that rely on Machine Learning (ML) for predictive modelling,  may suffer from the \textit{cold-start} problem: supervised models work well but, initially, there are no labels, which are costly or slow to obtain. This problem is even worse in imbalanced data scenarios, where labels of the positive class take longer to accumulate. We propose an Active Learning (AL) system for datasets with orders of magnitude of class imbalance, in a cold start streaming scenario. We present a computationally efficient Outlier-based Discriminative AL approach (ODAL) and design a novel 3-stage sequence of AL labeling policies where ODAL is used as warm-up. Then, we perform empirical studies in four real world datasets, with various magnitudes of class imbalance. The results show that our method can more quickly reach a high performance model than standard AL policies without ODAL warm-up. Its observed gains over random sampling can reach 80\% and be competitive with policies with an unlimited annotation budget or additional historical data (using just 2\% to 10\% of the labels).

\end{abstract}




\maketitle

\section{Introduction}\label{sec:introduction}

\begin{figure}[t]
    \centering
    \includegraphics[width=\linewidth]{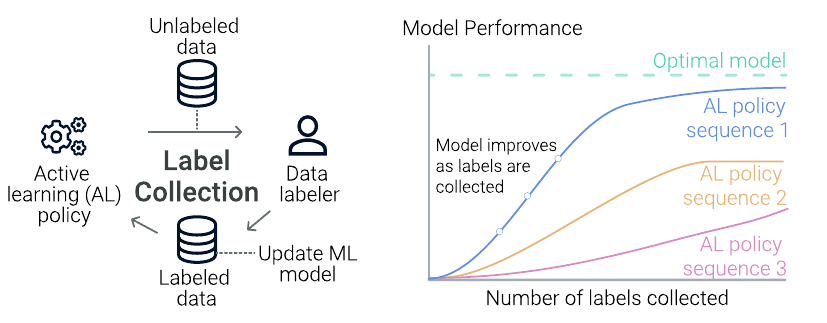}\vspace{-5pt}
    \caption{Schematics of the AL loop (left) and model performance evolution (right) depending on the AL policy.\vspace{-10pt}}
    \label{fig:AL_schematics}
\end{figure}

Training high performance supervised Machine Learning (ML) models is currently an essential and widespread task in the digital domain, where vast amounts of data are generated daily in numerous applications (e.g., finance, entertainment or consumer goods services).
Those models are often central in decisions that enhance system efficiency, user experience or even safety and they rely heavily on collecting high quality labeled data. In many use cases, labeled data is initially absent (\textit{cold start}) and expensive to collect, often requiring human annotation under a limited budget, while it is common that the system collects large amounts of unlabeled data. Thus, as data arrives and accumulates in the system,
it becomes essential to select the most informative samples for labeling to quickly be able to train a high performance ML model -- this is the goal of Active Learning (AL), 
as represented in Figure~\ref{fig:AL_schematics}.

In this paper, we propose an AL-based annotation method for real-time data streams with a large class imbalance, to train a high performance model in a \textit{cold start} scenario. We perform an extensive empirical study using real world datasets of credit card transactions where the ML task is to detect fraudulent transactions.
AL is especially relevant in this domain, since there is, usually, a considerable delay between the fraudulent event and the collection of the label (e.g., through client complaints or reports from financial institutions) unless a human analyst is consulted. 
We test well known AL policies, as well as our proposed sequences of policies that are especially designed for imbalanced datasets, to achieve a high performance, with reduced variance, in few iterations. 
Our main contributions are:
\begin{itemize}
\item A new computationally efficient approach to the Discriminative Active Learning method~\cite{discriminative_AL_DBLP:journals/corr/abs-1907-06347} named Outlier based Discriminative Active Learning (ODAL) -- Section~\ref{subsubsec:warm-up-policies-odal}.
\item Two variations of uncertainty sampling policies using an epistemic uncertainty measure, as well as a measure based on the fraud rate percentile -- Section~\ref{subsubsec:proposed_hot_policies}.
\item A 3-stage sequence of policies suited to highly imbalanced datasets, using ODAL in the second stage (\textit{warm-up} policy) -- Section~\ref{subsec:policy_sequences}. 
\item An extensive set of experiments to identify the best AL setup for fraud detection on four real world datasets  -- Section~\ref{sec:results}.
\end{itemize}

\section{Methods}\label{sec:methods}
In this section we present the \textit{use case} we will study in our experiments together with a description of the methods tested. Our basic problem consists of prioritizing instances from a growing unlabeled pool of data in a streaming environment for labeling (pool based sampling~\cite{settles2009active}) using AL.
The number of instances in one labeling query request (henceforth \textit{query}), i.e., the \textit{batch size}, is a parameter that may influence how fast AL improves the ML model trained with the collected labels. 
We focus on a realistic streaming data \textit{cold start} scenario, where initially, there is no labeled or unlabeled data available but rapidly unlabeled data accumulates. In particular, when AL selects a batch of queries for labeling, at a given iteration, more unlabeled data will be available than on previous iterations.

\begin{figure}
    \begin{center}
        \includegraphics[width=0.95\linewidth]{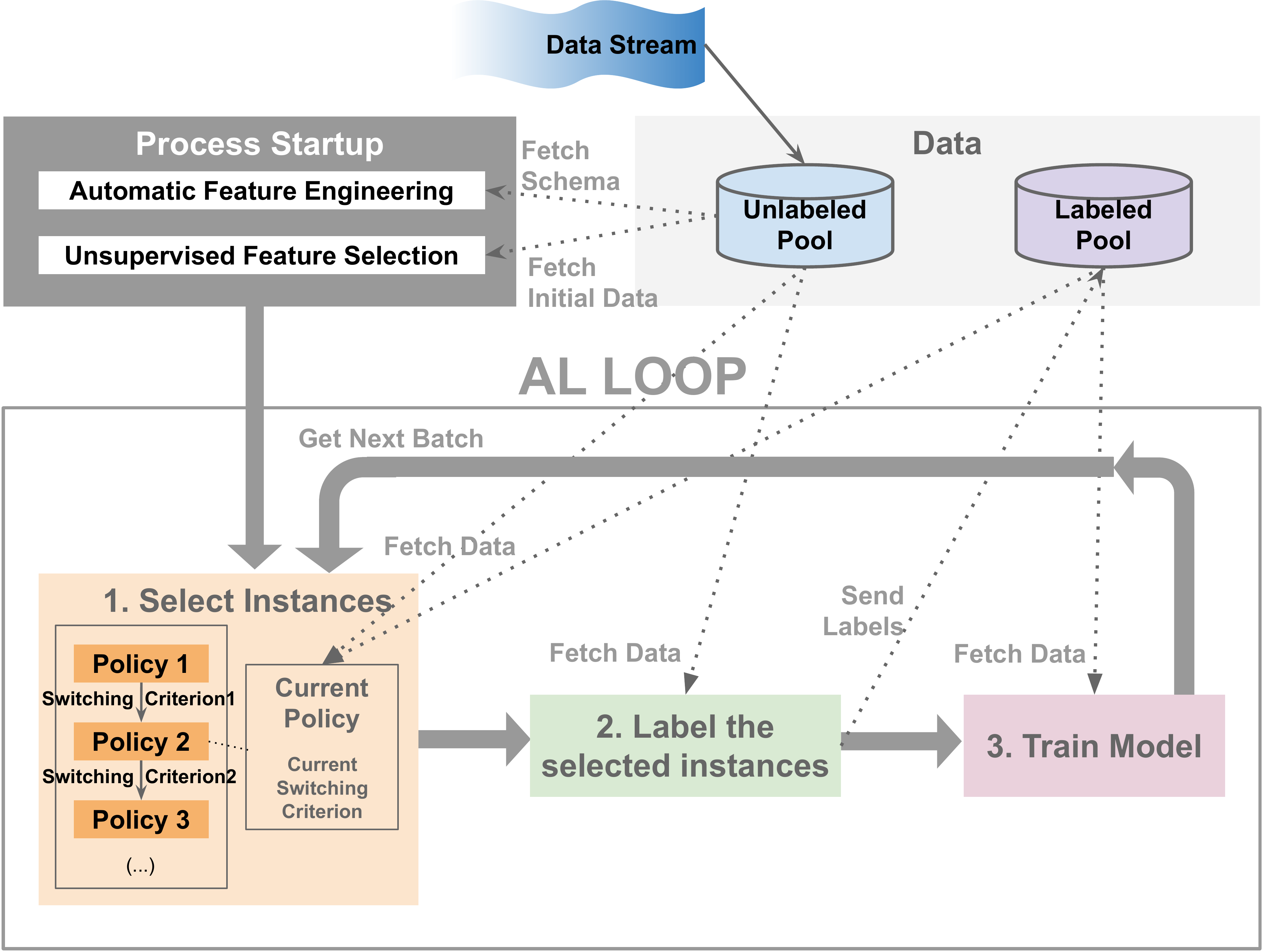}\vspace{-15pt}
    \end{center}
    \caption{\label{fig:system-architecture} Experimental framework architecture overview.\vspace{-15pt}}
\end{figure}
With this streaming data setup in mind, we present an illustrative diagram of the architecture of our full framework in Figure~\ref{fig:system-architecture} that will serve as reference when explaining the methods we developed.
Its main components are:
\begin{itemize}
    \item \textbf{Data Components}: This contains a \textit{Data Stream} collecting events in real time and storing them in the \textit{Unlabeled pool}.
    The \textit{Labeled pool} stores labeled data. Both pools start empty.
    \item \textbf{Process Startup}: This is responsible for training pre-processing pipelines, enriching the raw incoming data stream with features and applying feature selection and/or dimensionality reduction. 
    \item \textbf{AL loop}: This iteratively collects labels and trains the model.
    At each step the \textit{Data} is accessed and manipulated as follows:
    \begin{enumerate}
        \item \textit{Select Instances}: A batch of unlabeled events is selected by the currently active policy for querying.
        An arbitrary sequence of AL policies chained together with switching criteria is possible (left of block~1), though in our experiments we only consider up to 3-stage sequences.

        \item \textit{Label Instances}: Here we simply move the instances selected for labeling from the \textit{Unlabeled} to the \textit{Labeled} pool and reveal their label (our data sources contain the true label). In a live system, analysts would provide the labels.
        \item \textit{Train Model}: The labeled data is used to train and evaluate the ML model.
        We continuously iterate this loop up to a maximum fixed duration (e.g., until a 
        fixed number of labels is collected). 
        Since we use historical data to simulate the data stream, we can evaluate the sequence of AL models, obtained while iterating (Figure~\ref{fig:AL_schematics}), on a separate test set -- see Section~\ref{sec:experiments}.
    \end{enumerate}
\end{itemize}

\subsection{Startup and preprocessing}\label{subsec:preprocessing}

In a \textit{cold start} scenario we may not know, in advance, which features are useful to predict the target. 
Thus, we apply a minimal preprocessing depending only on the schema of the data fields collected by the system.
The transformations applied are now described.

\textit{Automatic Feature Engineering:}
We use Feedzai's AutoML tool~\cite{automl_feedzai_patent}, which generates a feature engineering plan based only on raw data fields.
This only requires a file specifying, e.g., grouping entities, numerical fields, or the semantics of fields to be used in pre-defined types of feature engineering operations, together with a specification of window durations to compute profile feature aggregations (e.g., count of transactions per card in the last hour).

\textit{Unsupervised Feature Selection:}
From both a data science and system performance perspective it is useful to discard redundant or noise features produced by AutoML.
We use \textit{Principal Component Analysis} (PCA)~\cite{pca_paper} to reduce the dimensionality of the numerical features space by selecting the top principal components that explain a fraction of the variance in the data. This method requires a sample of unlabeled data, which can be easily collected through an initial waiting period (e.g., we use one day in our experiments).

\subsection{Policies}\label{subsec:policies}
\begin{table}[t]
\begin{tabular}{l|l|l|l}
\hline
 & Cold stage & Warmup stage & Hot stage \\ \hline \hline
\multicolumn{1}{l|}{Baseline} & QueryAll & -- & -- \\ \hline
\multicolumn{1}{l|}{\multirow{2}{*}{1-stage}} & OutlierDetect & -- & -- \\
\multicolumn{1}{l|}{} & Random & -- & -- \\ \hline
\multicolumn{1}{l|}{\multirow{4}{*}{2-stage}} & Random & -- & Unc. (entropy) \\
\multicolumn{1}{l|}{} & Random & -- & ODAL \\
\multicolumn{1}{l|}{} & Random & -- & EMC \\
\multicolumn{1}{l|}{} & Random & -- & QBC \\ \hline
\multicolumn{1}{l|}{\multirow{5}{*}{3-stage}} & Random & ODAL & Unc. (entropy) \\
\multicolumn{1}{l|}{} & Random & ODAL & Unc. (epistemic) \\
\multicolumn{1}{l|}{} & Random & ODAL & Unc. (percentile) \\
\multicolumn{1}{l|}{} & Random & ODAL & QBC \\
\multicolumn{1}{l|}{} & Random & ODAL & EMC \\ \hline
\end{tabular}\vspace{-5pt}
\caption{\label{tab:policies} Policy sequences tested in experiments. \vspace{-20pt}}
\end{table}

The central ingredient in an AL based annotation system is the policy determining which instances are the most relevant to label.
We categorize the types of policies to mirror our three-stage strategy to efficiently train a model from \textit{cold start}:
\begin{enumerate}
    \item \textbf{Cold policies} (unsupervised): In a first stage, while no labeled data is available, a method is used to select the first instances for labeling before supervised AL can start  -- Section~\ref{subsubsec:cold-policies_methods}.
    \item \textbf{Warm-up policies}: After some labels are collected, there may be a transient period with only labels of a given type available (e.g., only negative class for binary classification) or too few labels to train a supervised policy -- Section~\ref{subsubsec:warm-up-policies-odal}.
    \item \textbf{Hot policies} (supervised): These are the most common, and they make full use of the collected labels to differentiate classes and select the best instances to query -- Section~\ref{subsubsec:proposed_hot_policies}.
\end{enumerate}
Next we discuss our choices for each stage. The combinations used in our experiments are summarized in Table~\ref{tab:policies}.

\subsubsection{Cold policies:}\label{subsubsec:cold-policies_methods}
AL studies in the literature often assume that a labeled pool is available to start the AL process.
However, in many real world scenarios one may be faced with a system that has just been deployed and contains no labeled data~\cite{pmlr-v32-houlsby14}.
Then, the initial sampling can only be guided by unlabeled instances.
The simplest choice is to randomly sample an initial batch of instances -- \textit{Random Policy}.
Another simple option is to use an unsupervised learning method to build a representation of the unlabeled data and select outliers -- \textit{Outlier Detection Policy} -- which is useful
if one or more of the classes behave as outliers.
We test, as 1-stage baselines (Table~\ref{tab:policies}), the \textit{Random} policy as well as an \textit{Outlier Detection} Policy consisting of an isolation forest~\cite{ISF_paper_4781136} trained on the unlabeled pool. The isolation score is used to rank the unlabeled transactions from most outlier-like (to query) to most inlier-like. Experiments using this method will be identified with the tag \textit{OutlierDetect}. 
\textit{Cold} policies are also baselines for AL if used alone (i.e., one-stage sequence experiments).

\subsubsection{Warm-up policies \& ODAL}
\label{subsubsec:warm-up-policies-odal}
Regarding \textit{warm-up} we propose a new method, \textit{Outlier Discriminative Active Learning} (ODAL),
that provides a computationally lighter approach to
\textit{Discriminative Active Learning (DAL)}
\label{subsubsec:discriminative-active-learning}
\cite{discriminative_AL_DBLP:journals/corr/abs-1907-06347}. The latter is based on the principle that a good labeled pool should be difficult to discriminate from the unlabeled pool.
In this approach, a binary classification model is fit to discriminate between pools, then the unlabeled pool is scored and instances that are easy to discriminate from the labeled pool are queried.
This can be computationally heavy because it always trains on all available data (labeled and unlabeled).
In ODAL we propose, instead, to train an outlier detection algorithm on the labeled pool, and use the obtained model to score the unlabeled pool and find the greatest outliers relative to the labeled pool. Those instances are then queried for labeling.
In typical AL scenarios the labeled pool is much smaller than the unlabeled pool, so ODAL can be trained on the labeled pool only, in contrast with DAL. Another advantage of ODAL over DAL is observed by expanding the DAL score, $p(0|x)$, for an instance with features $x$ to be in the unlabeled pool (0) using Bayes theorem (assuming a probabilistic discriminator):
\begin{equation}
    p(0|x) = \dfrac{p(x|0)p(0)}{p(x|0)p(0)+p(x|1)p(1)} = \left(1+\frac{p(x|1)p(1)}{p(x|0)p(0)}\right)^{-1}\;. \label{eq:ODALvsDAL}
\end{equation} 
Here $p(x|0)$, $p(x|1)$ are, respectively, the distributions of the unlabeled and labeled pools and $p(1)=1-p(0)$ is the fraction of labeled data. 
In Eq.~\eqref{eq:ODALvsDAL} we see that the DAL score is high for instances with a low density ratio, $p(x|1)/p(x|0)$, between the labeled and unlabeled pool, which may not be desirable if the labeled pool is missing examples in lower density regions of the unlabeled pool. On the other hand ODAL only models $p(x|1)$ so it favours, by design, that the instances to be selected are not well represented in the labeled pool regardless of how well they are represented in the unlabeled pool. For problems with a large class imbalance this is especially important. 
Thus, ODAL is both computationally feasible for our large scale experiments and less biased by the unlabeled data distribution.
We will see in Section~\ref{sec:results}, that ODAL warm-up adds an early boost to the learning curves in imbalanced datasets.
In the experiments, we will use an isolation forest outlier detection algorithm. 
Thus, the labeled pool instances will be ranked by isolation score and the ones ranking high are selected for labeling.

Within warm-up policies, there is another class of methods denoted \textit{Density-weighted} that
aim to select instances to cover well the most dense areas of the data distribution \cite{density_based_fujii-etal-1998-selective, density_based_cluster10.1145/1015330.1015349, density_based_cluster10.1007/978-3-540-71496-5_24}.
These methods tend to be heavier and harder to implement in streaming, because the unlabeled pool may grow and its distribution may drift in real-time, so we leave them out of our experiments.

\subsubsection{Hot policies}
\label{subsubsec:proposed_hot_policies}

We now describe the supervised policies.

\textit{Uncertainty Sampling:}
This is the most common active learning technique, originally discussed by Lewis and Gale~\cite{uncertainty_sampling_original_10.5555/188490.188495}.
It trains a machine learning model on each AL iteration using the labeled pool instances.
Then the unlabeled pool is scored and the instances are ranked by a measure of uncertainty related to the distance to the classification boundary.
Instances closer to the classification boundary are assumed to be more likely to improve the model.
A common criterion is to select instances with the highest expected entropy over the possible class labels given the model scores as the probabilities.
For binary classification those instances have scores closest to 0.5.
This method assumes scores that are well calibrated probabilities, which may not hold. Despite studies showing that it is an efficient AL uncertainty measure (\cite{review_YANG2018401} and references therein), the calibration assumption may not work for many algorithms and it can be especially bad for high class imbalance~\cite{calibration_pozzolo_7376606}.

We introduce an alternative for binary classification, that does not require score calibration.
We first observe that the score function of most ML algorithms is a monotonic function of the class posterior probability.
Thus we still expect that instances with higher scores will have a higher probability of being positive.
Given a sample of data, the classification boundary can be equivalently characterized by a score percentile, i.e., a position in the sorted set of scores.
We then note that the percentile of the classification boundary, for a perfect classifier that knows the labels would be equal to the negative class rate.
This motivates an alternative uncertainty criterion, where the uncertain instances are 
the ones closest to the estimated negative class rate boundary. 
In the experiments, we will show results with the classic \textit{entropy} criterion (denoted \textit{Unc. (entropy)}), as well as with our \textit{fraud percentile} criterion (denoted \textit{Unc. (percentile)}).

\textit{Expected variance Reduction and Epistemic Uncertainty:}
The expected variance reduction method estimates the variance of the model predictions~\cite{variance_reduction_COHN19961071}.
This is tightly related to the notion of epistemic uncertainty discussed in the literature~\cite{rf_uncertainty_10.1007/978-3-030-44584-3_35}.
Epistemic uncertainty is the reducible part of the total uncertainty composed of i) the model uncertainty (or bias)
, plus ii) the approximation uncertainty (variance). 
The remaining uncertainty (also know as aleatoric) is intrinsic to the data generating process and can never be removed.
The standard uncertainty sampling criterion using the entropy of the model scores is precisely the total uncertainty criterion.
The epistemic uncertainty, being the difference between the total and aleatoric uncertainty, may give a better measure of uncertainty for AL, because it is only sensitive to the reducible components.
Though it still contains the uncertainty from the bias, it can be more tractable than variance estimates, which often rely on analytic expressions assuming differentiability.
In our analysis, we train models using a random forest classifier. This is non-differentiable but it offers a convenient way of controlling regularization, using a large number of shallow trees, which is important to train on small labeled pools. 
The epistemic uncertainty for random forests is estimated from the outputs of each tree, ~\cite{rf_uncertainty_10.1007/978-3-030-44584-3_35}.
In our experimens, we denote the epistemic uncertainty criterion by \textit{Unc. (epistemic)}.

\textit{Query By Committee (QBC):}
Query by committee~\cite{qbc_10.1145/130385.130417}, is a simple but potentially computationally heavier method that combines knowledge from an ensemble of ML models, chosen by the user, where each model in the ensemble is trained on the labeled data pool and used to score the unlabeled data.
A measure of disagreement among the models 
is computed for each instance based on the model scores. Instances rank higher for higher disagreement.
Often, it also assumes that the scores are well calibrated probabilities. Thus, here we introduce an alternative measure of disagreement, among the models in the committee, that is insensitive to whether or not the scores output by each model are calibrated as probabilities.
This can be important if the committee contains a mixture of models with and without a probabilistic outcome.
For each model in the committee, we rank the unlabeled instances by descending model score and compute the average pairwise absolute difference of ranks between any two models.
Instances on which the models disagree will have very different rankings across models. 

\textit{Expected Model Change (EMC):}
The basic principle of this method \cite{expected_model_change_NIPS2007_3252} is to query the instance that is expected to change the model the most.
We use the simplest approach in the literature where: i) a gradient-based classifier is trained on the labeled data pool, ii) for each unlabeled instance, the expected gradient norm
for the given instance is computed assuming that the model parameters are near an optimum of the model's loss function and, iii)  the unlabeled pool instances are ranked so that instances with larger expected gradient are prioritized.
A related method that is often impractical is \textit{Expected Error Reduction}~\cite{expected_error_reduction_10.5555/645530.655646}, which requires retraining the model for all label assignments for each possible query. 

\subsubsection{Policy Sequences}
\label{subsec:policy_sequences}

In addition to the AL policy sequences displayed in Table~\ref{tab:policies} we also add a baseline denoted \textit{QueryAll} corresponding to unbounded labeling resources where all incoming transactions are labeled. 
In the 2-stage sequences we switch policies after we have at least one label from each class. Regarding the 3-stage sequences, the same criterion is used to switch between \textit{warm-up} and \textit{hot} policy, however, the switch to the \textit{warm-up} policy from the \textit{cold} policy is done after the first batch is collected with the \textit{cold} policy, to start exploiting ODAL immediately.

\section{Experiments}\label{sec:experiments}

In this section we present results of experiments with several real world credit card fraud datasets.

\subsection{Policy Parameters}
\label{subsec:policy_parameters}

We make the following choices for the various policies in the experiments. In all policies that require an isolation forest we use the \textsc{scikit-learn}~\cite{scikit-learn} implementation with 100 trees, using all features to grow each tree, and a maximum number of samples per tree  which is the minimum between 256 and the total number of samples. For \textit{QBC} we use a committee with: a Random Forest with 100 trees and maximum depths of 3, an L2 regularized Logistic Regression, a Gaussian naive Bayes classifier, and a Gradient Boosting Classifier with 100 estimators.\footnote{We use \textsc{scikit-learn}~\cite{scikit-learn} implementations for all mentioned ML models unless stated otherwise. For the unspecified hyper-parameters we use the library defaults.} For \textit{EMC} we use a logistic regression with L2 regularization.

As for the batch size, we use 100 for all policies. In preliminary experiments with smaller batch sizes we did not see substantial improvements, while larger sizes degrade the results.

\subsection{Data preparation}
\label{subsec:data-preparation}
We cover several representative use cases in the fraud detection domain, namely card issuing banks (Banking), platforms that process online payments for several merchants (Payment Processors) and single merchant online platforms (Merchants). 

\begin{table}
\begin{tabular}{r| r r}
\hline
Dataset & Positive class rate & Sampling fraction \\ \hline \hline  
Bank 1  &  $\sim 10^{-4}$                         & 11.0 \%                                \\
Bank 2  &  $\sim 10^{-3}$                       & 3.0 \%                             \\
Payment Processor    &  $\sim 10^{-2}$                         & 2.5 \%                              \\
Merchant    & $\sim 10^{-2}$                       & 100.0 \%                            \\
\hline
\end{tabular}
\caption{\label{tab:stats_datasets} \textit{Dataset properties:}
Due to privacy reasons we do not provide further details
(see detailed description in text).\vspace{-26pt}}
\end{table}
In Table~\ref{tab:stats_datasets} we provide some properties of each data set, which contain fraudulent (positive) and legitimate (negative) transactions. The fraud rates span several orders of magnitude, from an extremely large imbalance (Bank~1), to moderate imbalances of a few percent. 

The datasets contain raw fields collected when transactions arrived to a fraud detection system in real-time, including the monetary amount of the transaction, the timestamp of the event, several identifiers (e.g., card ID), categorical fields
and the fraud label.

The volume of transactions varies across datasets from a few millions to several hundreds of million per year.
To speed up our experiments, we applied undersampling to reduce the volume to a manageable (and similar) level for all datasets. This allowed us to scale up our experiments to cover many different types of policies and to perform a more extensive Temporal Cross Validation (TCV) over a longer period. We applied the sampling before feature engineering to speed up the preprocessing. Fraudulent and non-fraudulent card id entities were randomly sampled separately with the sampling rate indicated in Table~\ref{tab:stats_datasets} (this preserves the fraud rate) and all transactions were kept for each sampled card id. This keeps complete card histories, allowing to compute sliding window profiles that are important to characterize the event \cite{rnns_feedzai_10.1145/3394486.3403361}.

We applied automatic feature engineering, which generated between 600 and 800 features depending on the dataset -- see Section~\ref{subsec:preprocessing}. The categorical fields were encoded both with ordinal and frequency encoding and standardized to zero mean and unit variance similarly to other numerical features. The remaining pre-processing is scenario specific -- details provided in Section~\ref{sec:experimental_setup}.

\subsection{Experimental Setup}\label{sec:experimental_setup}

In this section we describe details of the experimental setup that are common to all data sets. 
\begin{figure}
\begin{center}
 \includegraphics[width=0.98\linewidth]{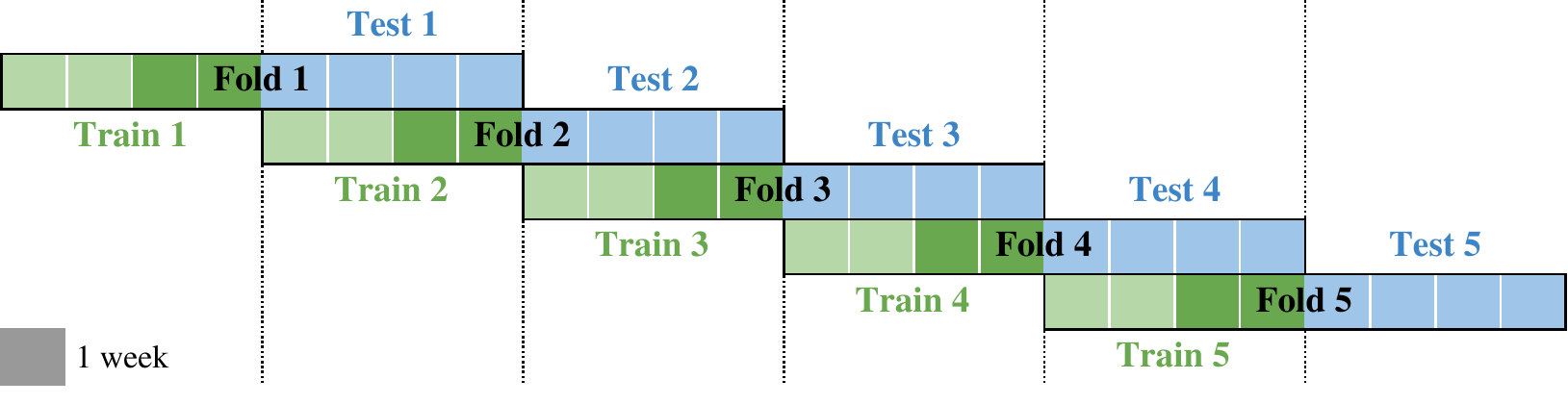}\vspace{-10pt} 
\end{center}
    \caption{\label{fig:time-folds} Time folds for the five simulation periods in the experiments (see detailed description in the text).
    \vspace{-10pt}
    }
\end{figure}

\subsubsection{Data slicing}

In Figure~\ref{fig:time-folds} we present a diagram of the various slices of data for any given data set. We define \textit{Folds}, which consist of 8 week periods (two pairs of 4 weeks). Within each fold, the first 4 weeks (green), are used for model training, whereas the following 4 weeks (blue), are for model evaluations. The \textit{Train} period is used differently according to the type of experimental run.
\subsubsection{AL scenario and baselines}
Here we describe the details of data preprocessing pipeline preparation and training periods.
\vspace{-3pt}

\subsubsection*{AL in streaming} This case mimics a scenario where the AL system is deployed for the first time in streaming without access to previous data. Since the goal is to collect labels quickly to obtain a good model, without waiting for labels to arrive by other means, applying AL is typically relevant for a few weeks.
Thus, we only reserve the two last weeks of the \textit{Train} periods (darker green: weeks~3 and 4) to sample data with AL for training (weeks 1 and 2 are used for the strong optimistic baseline discussed next). The \textit{Test} sets allows us to measure the model performance after the deployment of the last AL model. In practice, for most data sets we only use one week for AL training (except for Bank~1 which, due to the extreme class imbalance, needs a longer period for the performance to stabilize).\vspace{-3pt}

\subsubsection*{Optimistic Baseline} Here we train a strong model that has access to all \textit{Train} data and labels (weeks~1 to 4: light plus dark green). The goal is to obtain a ``best case scenario'' upper bound performance.

\subsubsection{Training procedure}

Each experiment (either AL or Optimistic Baseline) consists of 35 repetitions of the train-test procedure  with different pseudo-random number generator seeds. This allows us to assess the stability of the AL policies by observing the variance of our metrics. We choose 35 seeds as a good trade off between run time and a high chance of observing a wider range of values around the center of the distribution.
As displayed in Figure~\ref{fig:time-folds}, we repeat each experiment in 5 different folds (\textit{Train}+\textit{Test} pairs) to observe the robustness of the AL procedure against temporal variations.\vspace{-3pt}

\subsubsection*{Streaming AL Training}
In all AL experiments we include an initial waiting period of one day to simulate the collection of some unlabeled data to fit the pre-processing pipeline. This mimics a realistic scenario of deployment with no previous data. To reduce the number of numerical features generated by the AutoML pipeline (which may contain redundant information) we apply PCA on the numerical features. 
In preliminary experiments on Bank~2, we checked that about 90 features can explain $99\%$ of the data variance. Then we decided to fix 90 features after PCA for all data sets to keep the run time similar across experiments.

Observe that our pre-processing pipeline is trained on the first day of unlabeled data, and used to transform all future data arriving at the stream (\textit{Train} or \textit{Test} period). This is to mimic a day-1 system deployment. However, after day-1, the pipeline could be updated frequently but, for simplicity, we chose to fix it in our experiments.

For each run, several labeling iterations are processed after the waiting period of one day, according to the diagram of Figure~\ref{fig:system-architecture} -- see Section~\ref{sec:methods}. Therefore the unlabeled pool grows with time, as does the labeled pool during the AL training iterations, whose growth is indirectly controlled by the time assumed for the team of analysts to label each queried batch of events. 
Thus, if the team is, e.g., a single analyst taking one hour to review a batch, we assume that one hour of new data is inserted in the unlabeled pool after the batch is labeled.
For simplicity we use a fixed batch size and a fixed time to review corresponding to an overall review rate of 1000 events per day. The only exception is for Bank~1, where, due to the extreme class imbalance, we assumed twice the daily budget. 

Regarding the ML model to train on the AL labeled data, we chose a highly regularized Random Forest (RF) classifier from the \textsc{scikit-learn} library with a maximum tree depth of 3 and 200 trees (other hyper-parameters set to defaults).
We did a small study on Bank~2 on two time folds, where we either, i) varied the number of trees up to 1000, ii) reduced or increased the maximum depth, or iii) used other models with various different levels of regularization (Feed Forward Neural Networks, Support Vector Machines and Naïve Bayes). This confirmed the benefits of regularization. Despite improvements with 1000 trees, we chose 200 to speedup our simulations. \vspace{-3pt}

\subsubsection*{Optimistic Baseline Training}
Here we assume access to fully labeled data in the 4 weeks of the \textit{Train} period.
Additionally we apply a more robust training methodology. We train a RF classifier with 300 trees and a maximum depth of 20. For each of the 35 models (one per seed) we train 5 random configurations of hyper-parameters on the first 3 weeks and evaluate on week~4 to select the best configuration. The final configuration is re-fit on the 4 weeks.

For each model trained above, we also apply supervised feature selection. 
The fraction of features to use is a hyper-parameter to vary. In addition, we also vary the minimum number of samples in a leaf node, a binary parameter (to use class weights or not), and the complexity parameter for minimal cost-complexity pruning. 

\subsubsection{Evaluation metrics}
\label{subsubsec:evaluation_metrics}

We now discuss the performance metrics used to measure the quality of a single AL experiment, as well as to aggregate and summarize an experiment to compare runs. \vspace{-3pt}

\subsubsection*{Learning curves}
A single AL experiment, consists of several iterations where the labeled pool grows, and a sequence of models that can be evaluated on the \textit{Test} set are trained. Given a performance metric (e.g., recall at a fixed false positive rate), we obtain a learning curve where the metric usually improves during the simulation. Since we run 35 simulations, we obtain a distribution of learning curves, which we will visualize as percentile band plots 
in Section~\ref{sec:results}.

Since we run hundreds of experiments to test different policies, datasets and time periods, it is not feasible to observe all learning curves. Therefore we now define three aggregations to summarize each set of learning curves and be able to interpret the results.\vspace{-3pt}

\subsubsection*{Learning curves rise} To summarise how quickly the learning curves rise throughout the iterations (see, e.g., Figure~\ref{fig:learning_curves_bank1}), we compute the Area Under the percentile 50 learning curve (Area P50), defined as the curve tracing the median performance (over the 35 seeds) on each iteration. In addition, we normalize it by the area under the median optimistic baseline, which is the horizontal line corresponding to the median performance of the optimistic model (denoted by \textit{Norm Area P50}). This allows us to compare folds relative to their optimistic baseline, while correcting for temporal drift unrelated to AL that also shifts that baseline.\vspace{-3pt}

\subsubsection*{Learning curves variability} 
To measure the variance of the learning curves (a good policy will always rise fast for all seeds), we use the Area between the percentiles 10 and 90 (denoted \textit{Var} in the results). This is also normalized by the optimistic baseline area.\vspace{-3pt}

\subsubsection*{Quality of the final AL model}
This is defined as the median performance of the final AL model normalized by the performance of the optimistic baseline (we denote it as \textit{Norm Final P50}).  

\section{Results}\label{sec:results}

We now present results of the AL experiments
for the various 
datasets
focusing first on the most imbalanced.
\begin{table*}[tb]
    \vspace{-10pt}
    \hspace{10pt}\includegraphics[width=0.82\linewidth]{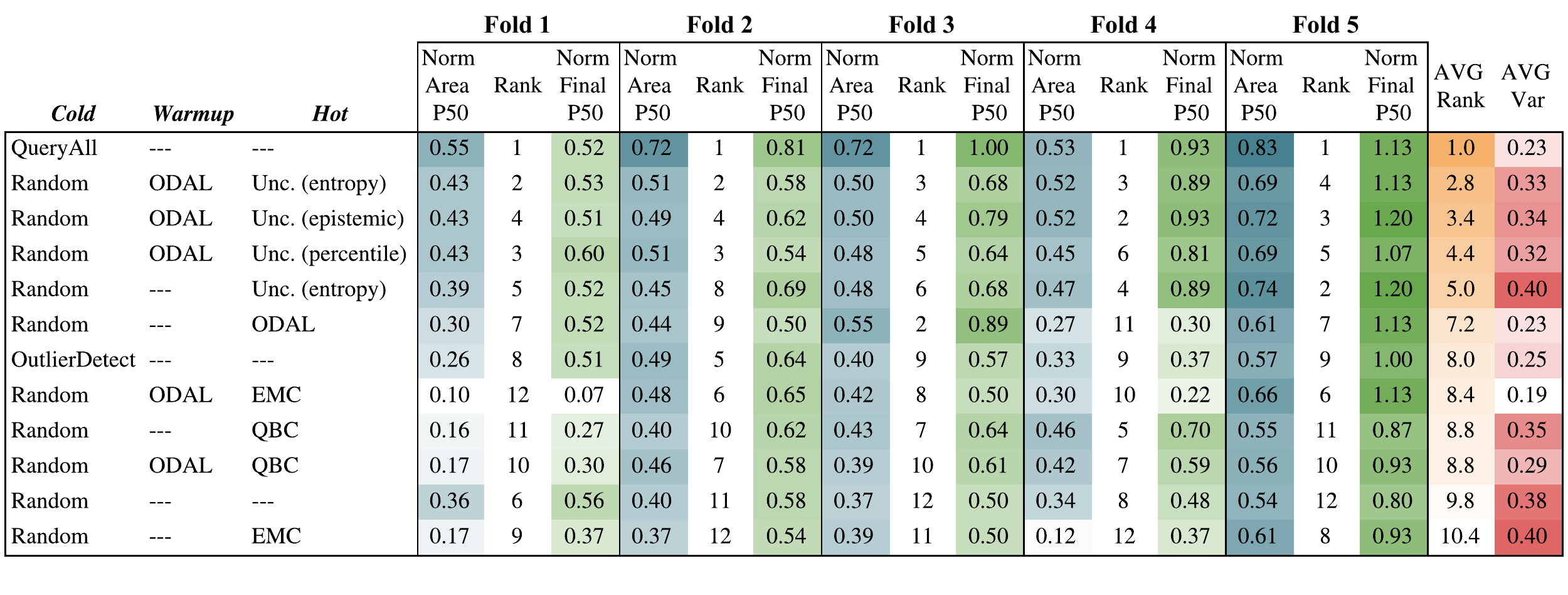}\vspace{-20pt}
    \caption{\label{tab:bank1_rankings}\textit{Bank 1 rankings of AL policies using various folds} (see also detailed description in the text).\vspace{-15pt}
    }
\end{table*}
\begin{figure*}[tb]
    \includegraphics[trim={3cm 0 0cm 0},clip,height=0.252\textwidth]{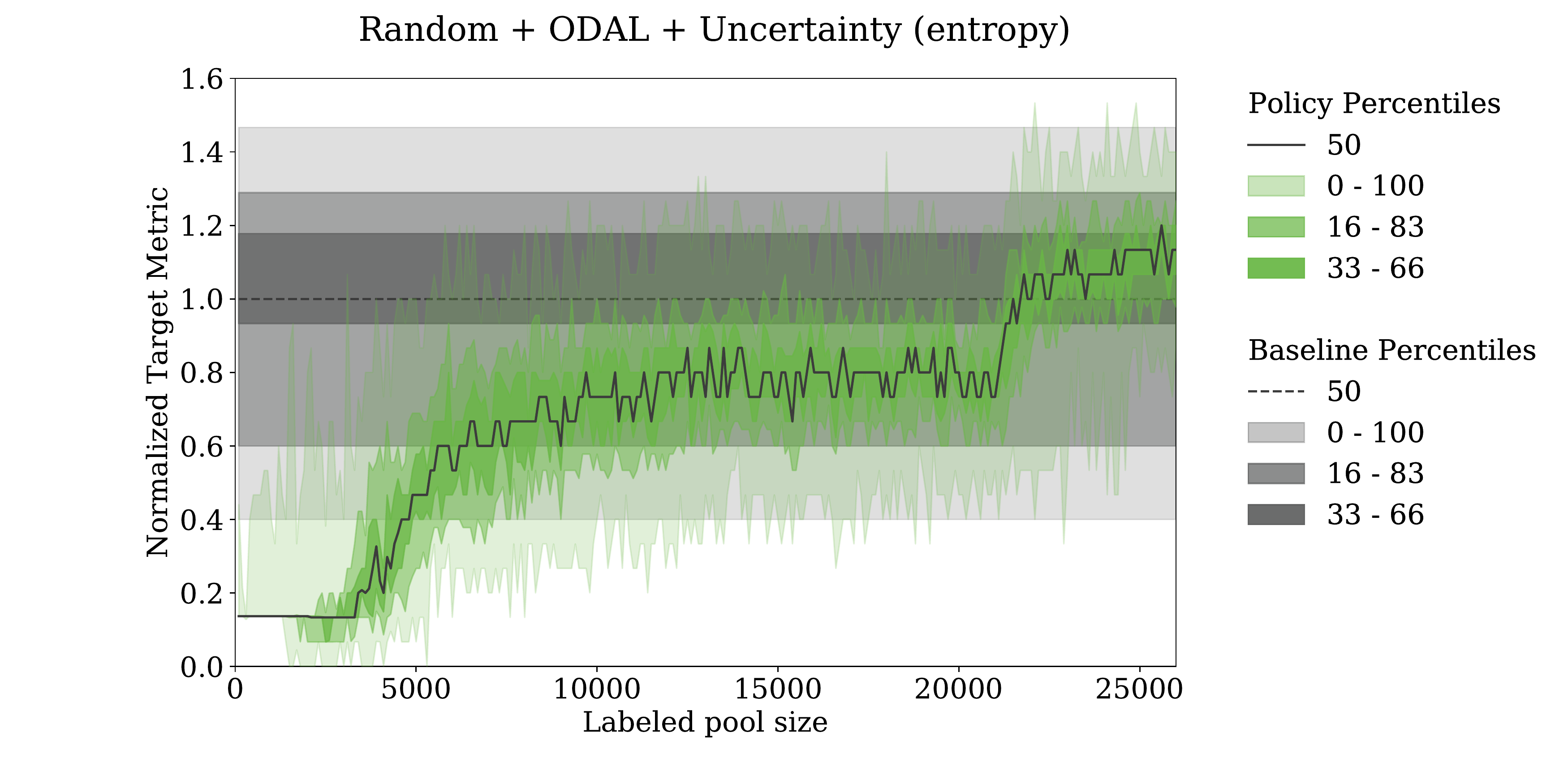}\hspace{0.05\linewidth}\includegraphics[trim={3cm 0 8cm 0},clip,height=0.252\textwidth]{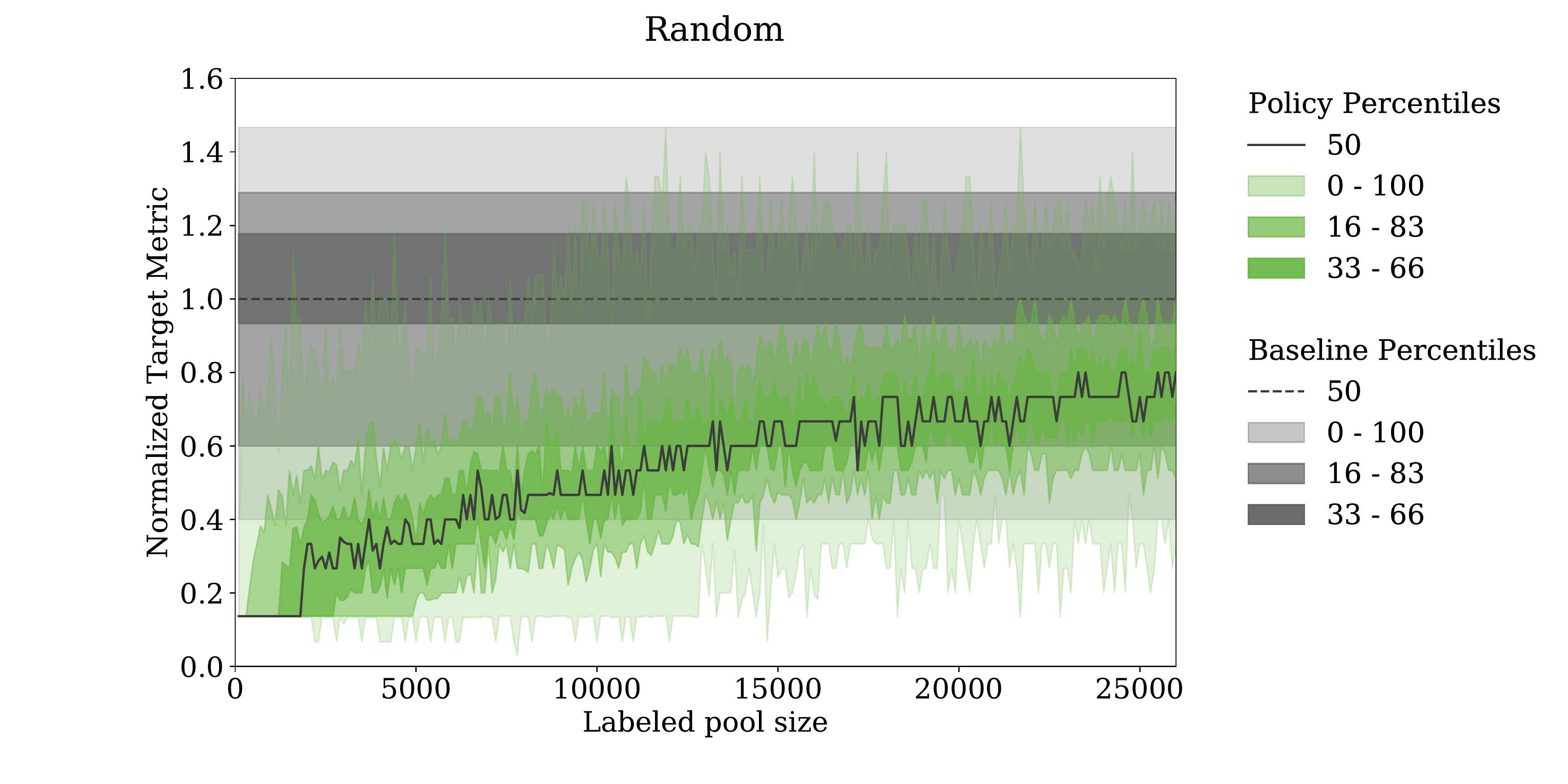}\vspace{-10pt}
    \caption{\label{fig:learning_curves_bank1} \textit{Learning curves distribution for Bank~1 in the best fold (5)}: The best sequence of policies (left panel green bands), 
    and the \textit{Random} policy (right panel green band), normalized by the percentile~50 of the optimistic baseline (gray bands).\vspace{-5pt}
    }
\end{figure*} 
In Table~\ref{tab:bank1_rankings}, we show a summary of metrics for the five folds and all policies for Bank~1.  
Each row displays values for a specific policy sequence (1-stage, with only a cold policy, and 2-stage, without warm-up -- see dashed lines).  We have five groups of columns (one per \textit{Fold}) with three metrics each (see Section~\ref{subsubsec:evaluation_metrics}): i) the normalized area under percentile~50 (Norm~Area~P50, blue density scale), ii) the ranking of the policy for the fold according to Norm~Area~P50 (center), and  iii) the percentile~50 of the final normalized AL model performance (Norm Final P50, green density scale).

The rightmost pair of columns in Table~\ref{tab:bank1_rankings} contains two metrics that summarize the five folds, namely the average of the ranks of each fold for each sequence (AVG Rank, orange density scale) and the average of the normalized area between percentiles 10 and 90 (AVG Var, red density scale). The former provides an overall measure of how fast the policy performance rises, whereas the latter of how noisy the policy is, for this dataset. The table rows are sorted by ascending AVG~Rank. Therefore policies that perform better on various folds are at the top. We choose to rank by Norm~Area~P50 rather than Norm~Final~P50 because it is more sensitive to how quickly the learning curves rise, which is critical in systems that need a good model to start acting as early as possible. Nevertheless, the final model performance is important to tells us how close we get to the optimistic baseline. We include 12 sequences specified on the left.
\textit{Random} and \textit{QueryAll} are baselines (Section~\ref{subsec:policy_sequences}).

Bank~1 is the most challenging dataset with an extremely large class imbalance. Therefore we doubled the daily review budget and trained in the full two weeks available for AL in the \textit{Train} period (see weeks 3 and 4 of each \textit{Fold} in Figure~\ref{fig:time-folds}). The best policies in Table~\ref{tab:bank1_rankings} outperform \textit{Random} by a large margin (close to doubling the performance in some cases). Furthermore, they are on par with the \textit{QueryAll} on folds 1, 4 and 5, both for the \textit{Area} metric and the \textit{Final} performance. In folds 2 and 3, although \textit{QueryAll} performs substantially better, the group of top performing AL policies, based on uncertainty sampling, continue to rank highly. 

Observe that, except for the rank, all the metrics have been normalized by the optimistic baseline, which is trained on extra data (full 4 weeks of the train period vs 2 weeks in Figure~\ref{fig:time-folds}) with supervised feature selection and hyper-parameter tuning. 
This additional data would not be available in a realistic production setting and the improved training is challenging for AL in streaming. This explains why most metrics are smaller than~1. The exception is \textit{Fold~5}, where Norm~Final~P50 is larger than~1 for various policies. 
This can be explained by observing the learning curves for \textit{Fold~5} in Figure~\ref{fig:learning_curves_bank1}, where we show the distribution of learning curves for the best AL policy (left) and the \textit{Random} policy (right)-- represented by the rising green bands. Three equally spaced percentile bands are included, together with a solid gray line that traces the median. The distribution of values for the optimistic baseline is represented in the horizontal gray bands. All values have been normalized by the percentile 50 of the optimistic baseline. In this fold we can see that the distribution of values for the training of the optimistic baseline is quite wide. Thus, despite being above~1, the final performance of the AL model for the best policy is still within the central part of the distribution.
Comparing left and right, we confirm that the 3-stages policy rises quick to high performance with a narrow variance. 

It is also important to note that 3-stage sequences, i.e., with ODAL \textit{warm-up}, tend to outperform the corresponding setups without ODAL, especially when paired with uncertainty based policies.

The overall conclusions, up to data set specific noise and some temporal drift effects, are confirmed for the other datasets.
Note that AL typically only uses 2\% to 10\% of the number of samples available to the optimistic baselines.
For other datasets we only present the policy rankings in Section~\ref{subsec:aggregation-over-datasets}, due to space constraints. 

\subsection{Aggregation over Datasets}
\label{subsec:aggregation-over-datasets}

In the previous section we discussed policy rankings and a pattern emerged: 3-stage sequences were the best performing policies, some 2-stage sequences also showed a good performance, and the rankings of the least performing policies were unstable across folds. 

A convenient way of aggregating this information, to provide a clearer picture of the overall rankings, is to average out the policy ranks over the studied datasets. This is displayed in Table~\ref{tab:overall_ranking}.
\begin{table}[tb]
    \begin{center}
        \includegraphics[width=\linewidth]{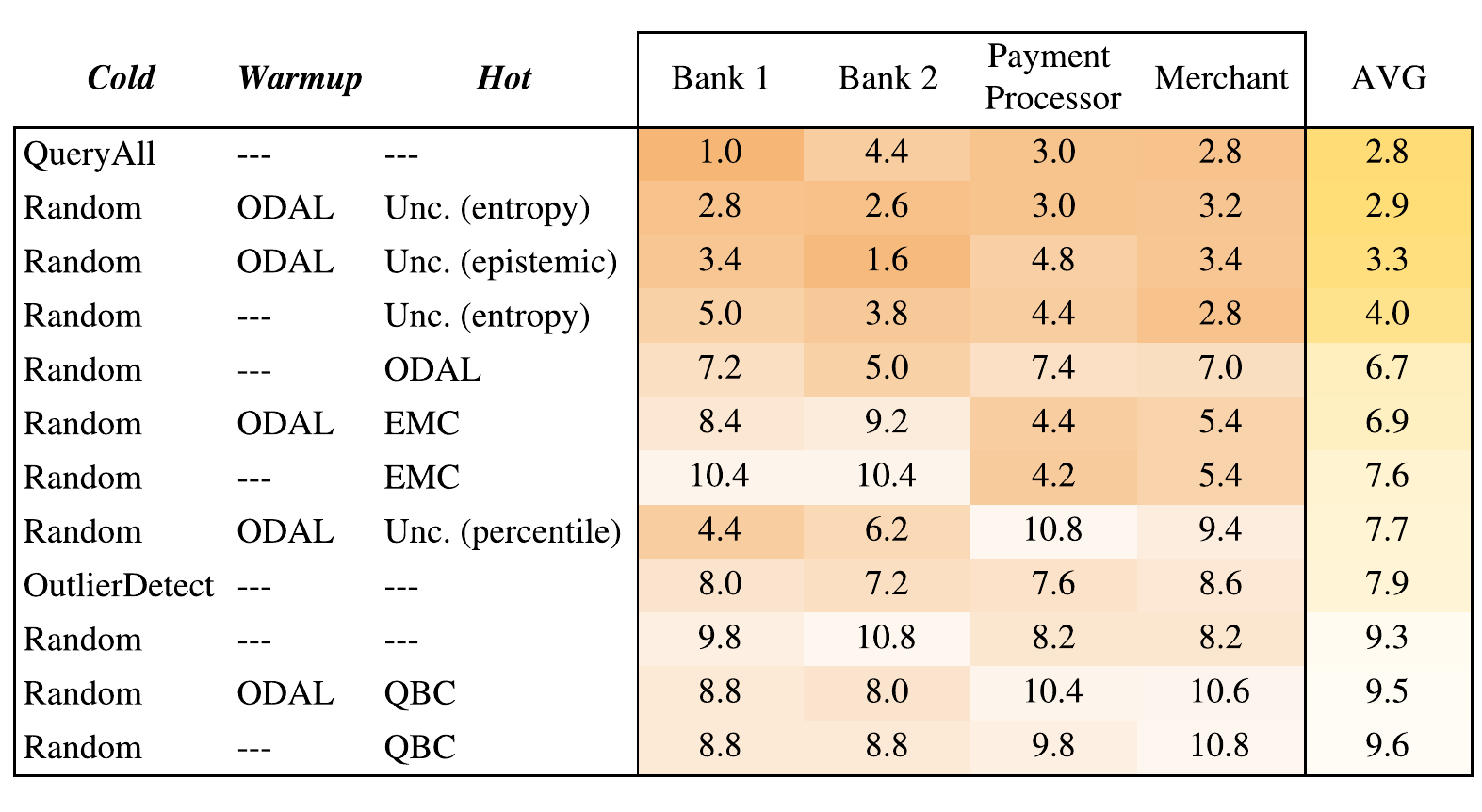}\vspace{-5pt}
    \end{center}
    \caption{\textit{Overall policy ranking}: Average ranks for each dataset (four central columns) and their overall average (right column). Rows are sorted by the AVG column.\vspace{-25pt}}
    \label{tab:overall_ranking}
\end{table}
As expected, overall, the \textit{QueryAll} policy ranks first, even though it is not always the top one for some datasets. The 3-stage policies based on entropy or epistemic uncertainty rank very close to it, which indicates that these are high quality AL policies. Regarding sequences with \textit{Expected Model Change} or the fraud percentile based \textit{Uncertainty} policy, despite ranking in the middle of the table, for some datasets they rank very low, so they are not very stable/consistent. On the other hand, the 2-stage policy with ODAL ranks between 5 and 7 across datasets, which reinforces its value as a stable \textit{warm-up} policy.  The \textit{Random} policy ranks low, as expected. \textit{QBC} also ranks low, but this may be due to our specific/simple choice of committee (a more detailed study is left to future work).
Another important observation 
is that all 3-stage policies rank higher than their 2-stages counterpart.
\begin{figure}[tb] 
\begin{center}
    \includegraphics[width=\linewidth]{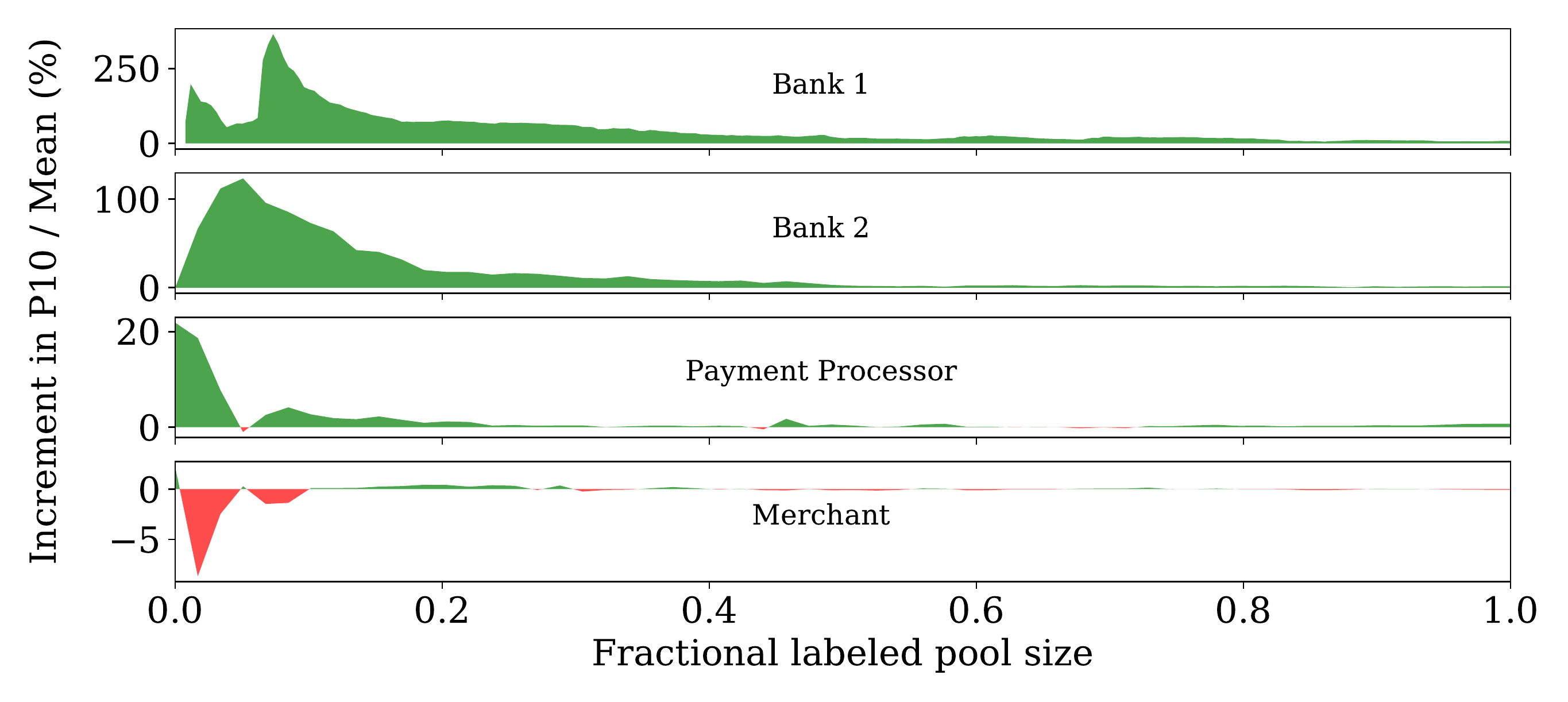}\vspace{-15pt}
\end{center}
  \caption{\textit{Boost in the number of positives sampled in 3-stages vs 2-stages for the entropy based uncertainty policy} (see detailed description in the text).\vspace{-15pt}
  }
  \label{fig:2vs3stage}
\end{figure} 
In Figure~\ref{fig:2vs3stage} we display a visualization that helps understanding this improvement for the entropy based uncertainty policy.  
On each row we present the average increase of sampled positives, over all folds, when adding ODAL as a warm-up policy. For each fold, the increase is the 10th percentile difference between the positives obtained with a 3-stage sequence and the corresponding 2-stage sequence, divided by the mean positives of the 2-stage sequence.
We can clearly observe that, for datasets with larger imbalances, including ODAL lifts up this low percentile 
considerably in early iterations (e.g., $\sim 3\times$ the mean value for \textit{Bank~1}).
The effect progressively 
disappears for
milder imbalances -- \textit{Merchant}.

\section{Related Work}
\label{sec:related-work}

Various AL methods have been proposed and surveyed in the literature in the last decades, ~\cite{settles2009active,review_YANG2018401}.
In our experiments we focused on the \textit{cold start} scenario with no historical data in a streaming environment where the unlabeled data pool grows.
This is in contrast with typical AL setups where the data source is static.  
Some studies have appeared in the literature discussing AL methods in a streaming data scenario~\cite{AL_streaming_10.1007/978-3-642-23808-6_39,AL_streaming_5440901,OAL_8910490, Carcillo2018StreamingAL,text_classification_modifiedQBC,OAL_drift,network_data_10.1145/2661829.2661981, JANARDAN2017804,partial_dishcarge_use_case_7245026,sentiment_analysis_KRANJC2015187}.
Notably, Carcillo et al.~\cite{Carcillo2018StreamingAL},  investigated several AL methods for a credit card fraud dataset. Instances were selected once a day with AL, according to a fixed budget.
In contrast, we consider scenarios where several small batches of instances are processed during the day to exploit the collected labels more frequently to update the AL policy, which is important to avoid the selection of many similar instances in one large batch.
Furthermore, we presented a detailed analysis of AL curves in the fraud domain, to provide a more complete understanding of its effectiveness for fraud detection, as well as investigated new policies that were not considered in Carcillo~et ~al.~\cite{Carcillo2018StreamingAL}. Finally, in reference~\cite{Carcillo2018StreamingAL} no analysis of AL curves was presented, nor of their variability, which is essential to observe the boost in ML model performance at early stages of the AL process. Other studies in our literature review, cited above, are either: i) focused on applying AL to address concept drift, or ii) not focused on highly imbalanced problems, or iii) not focused on dealing with the \textit{cold start} problem.

\section{Conclusions}\label{sec:conclusions}

We studied the problem of creating a small labeled dataset, with a limited budget of annotations by analysts, in a streaming environment, in a cold start scenario (no previously labeled data and little or no unlabeled data) for highly imbalanced datasets. We proposed an AL system adapted to these conditions and performed a detailed study on four real world credit card fraud detection datasets, covering three use cases with several orders of magnitude in class imbalances. We proposed various ingredients that proved essential, namely: i) ODAL, a computationally efficient version of discriminative active learning to quickly represent well the unlabeled pool in the labeled pool, relying only on the labeled pool features distribution, and ii) the combination of ODAL, as a warmup-policy, with other AL polices, in a 3-stage sequence to alleviate the cold start problem in highly imbalanced datasets where it may take a long time until some of the labels are found.
We also proposed two alternative uncertainty measures for the \textit{Uncertainty Sampling} policy -- epistemic uncertainty and the fraud percentile measure -- as well as an alternative measure of disagreement based on rank differences for \textit{Query By Committee}.

In Section~\ref{sec:experiments} we conducted detailed experimental studies, including optimistic baselines and 12 different policy sequences to be ranked. Our analysis showed that the best performing AL policies are 3-stage sequences with ODAL warm-up and Uncertainty Sampling as Hot policy (either entropy or epistemic). In particular, we showed that the ODAL warm-up boosts the learning curves in the earlier AL iterations. As a general rule, the final overall ranking shows that including ODAL warm-up before any \textit{Hot} policy boosts its learning curves, especially for large class imbalance. 
Furthermore, the best performing sequence is often as good as the \textit{QueryAll} policy, it has low variance learning curves, it is competitive with the optimistic baseline and substantially better than \textit{Random}. Our results show that the required amount of labeled examples, until the learning curve stabilizes, often ranges between $3\,000$ to $6\,000$ for mild to intermediate class imbalances, and a bit over $20\,000$ for extreme imbalances ($\sim 2\%$ to 10\% of the optimistic baseline data).

To conclude, we comment on some future directions.
In this study, we have simulated the analyst queries by using the real labels in the datasets. It would be interesting to perform experiments with real analysts in a live environment to see if the performance gains are confirmed.
Finally, we have not touched upon other possible problems and improvements that could be important in a real system. This includes the issue of evaluating the AL models online -- in our study we used an independent test set in the future of the train set for evaluation. Related to this, it would also be interesting to include online hyper-parameter tuning and model selection, as well as online supervised feature selection, instead of using a static set of features selected in an unsupervised way on the first day.

\begin{acks}
We thank Jacopo Bono for reviewing the manuscript. The project CAMELOT (reference POCI-01-0247-FEDER-045915) leading to this work is co-financed by the ERDF - European Regional Development Fund through the Operational Program for Competitiveness and Internationalisation - COMPETE 2020, the North Portugal Regional Operational Program - NORTE 2020 and by the Portuguese Foundation for Science and Technology - FCT under the CMU Portugal international partnership.
\end{acks}



\bibliographystyle{ACM-Reference-Format}
\bibliography{references}

\end{document}